# Kaldi+PDNN: Building DNN-based ASR Systems with Kaldi and PDNN

*Yajie Miao*

Language Technologies Institute, School of Computer Science, Carnegie Mellon University

The Kaldi[1] toolkit is becoming popular for constructing automated speech recognition (ASR) systems. Meanwhile, in recent years, deep neural networks (DNNs) have shown state-of-the-art performance on various ASR tasks. This document describes our recipes to implement fully-fledged DNN acoustic modeling using Kaldi and PDNN. PDNN is a lightweight deep learning toolkit developed under the Theano environment. Using these recipes, we can build up multiple systems including DNN hybrid systems, convolutional neural network (CNN) systems and bottleneck feature systems. These recipes are directly based on the Kaldi Switchboard 110-hour setup. However, adapting them to new datasets is easy to achieve.

## 1. Introduction

Deep neural networks (DNNs) have shown superior performance over the traditional state-of-the-art GMM-HMM on ASR tasks [1, 2]. When applied in ASR, DNNs have multiple hidden layers and model context-dependent states directly. In this paper, we present our Kaldi recipes to build DNN acoustic models using PDNN[2], a lightweight deep learning toolkit developed on top of Theano[3]. The rough idea behind these recipes is to build the initial GMM models with Kaldi, train DNNs with PDNN and finally load the DNN models back to Kaldi for further decoding or system building. More details about the pipeline can be found in Section 2. We elaborate on PDNN and individual recipes in Section 3 and 4. The recipes have the following key features:

A. Consistent with Kaldi
- The recipes are written in the Kaldi style and thus can be integrated with any existing recipes seamlessly.

B. Diverse Systems
- Multiple systems with different architectures; potentially useful for system combination.

C. Easy to Run and Modify
- The recipes are run in the "one-button" fashion, producing word error rates (WERs) at the end without any user intervention.
- All the DNN configurations (e.g., network structure, learning rate, output format, etc.) are visible when PDNN commands are called.

D. Convenient for Research
- The PDNN toolkit is open source and written in python. Experiments with new research ideas become easy and fast.

E. Open License
- All the recipes, including the PDNN toolkit, are licensed under Apache 2.0, one of the least restrictive licenses.

## 2. Overall Pipeline

The current release of Kaldi+PDNN consists of 4 DNN recipes. Fig. 1 shows the overall pipeline for these recipes. More information about individual recipes will be presented in Section 4.

The recipes start with initial GMM models which can be obtained after we run the existing Kaldi recipes, e.g., egs/swbd/s5b/run.sh for the Switchboard setup. Training and validation data for DNNs are generated from the GMM models by performing forced alignment. By default, our recipes use the speaker adaptive training (SAT) model.

Then, various networks are trained with PDNN and saved in the Kaldi-supported format. Some recipes require multiple networks to be trained. For instance, in the BNF+DNN recipe, we need to train the bottleneck feature network and the DNN on top of bottleneck features. Note that for DNN training, Fig. 1 is not showing other optional steps such as pre-training.

Finally, we load the saved networks back to Kaldi. If we are building hybrid systems, the trained networks can be used directly for decoding. Alternatively, if the networks have the bottleneck structure, then GMM systems, which are also referred to as tandem systems, can be further constructed over the bottleneck feature representations.

---

[1] http://kaldi.sourceforge.net/
[2] http://www.cs.cmu.edu/~ymiao/pdnntk.html
[3] http://deeplearning.net/software/theano

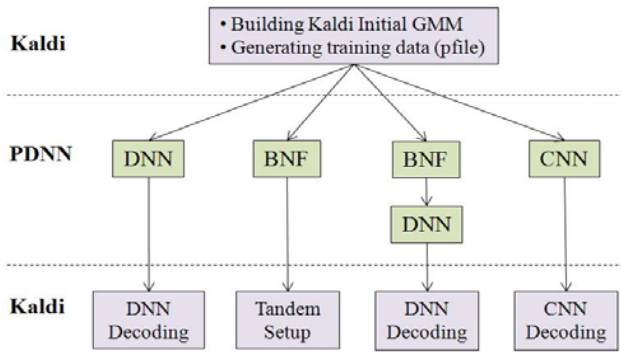

Figure 1. Overall Pipeline of our Kaldi+PDNN Recipes.

## 3. PDNN: Yet Another Python Toolkit for Deep Neural Networks

PDNN is a lightweight deep learning toolkit developed under the Theano environment. It takes advantage of two Theano's features: GPU utilization and gradient computation. Therefore, Theano needs to be installed properly on your computing machine, where a GPU card should be available to speed up DNN training. Check the PDNN website http://www.cs.cmu.edu/~ymiao/pdnntk for details about installation, configuration and usage.

Currently, the input data into PDNN have to be in the format of "PFile", the ICSI feature file archive format. Future releases will support more data formats. Table 1 demonstrates the PFile format and lists 3 example lines. Each row starts with the utterance and frame index, followed by the feature vector. The last field represents "class label" which in speech recognition means context-dependent state label for each frame. Note that frame index is based on each utterance, instead of being sequenced globally. You can specify the following arguments for PFile reading.

- partition. Due to the limitation of GPU memory, we have to split PFile data into partitions and load one partition to GPUs each time. The argument "partition" indicates the size of each partition.
- random. This argument specifies whether the examples in one partition are shuffled when they are loaded to GPUs.
- stream. If "stream" is true, then PDNN reads one partition of PFile data from local storage and loads this partition to GPUs. Otherwise, PDNN keeps the entire PFile data in the CPU memory and feeds data to GPUs partition by partition.

Table 1. PFile format and example lines.

| Utterance Index | Frame Index | Feature Vector | Class Label |
|---|---|---|---|
| 0 | 0 | [0.2, 0.3, 0.5, 1.4, 1.8, 2.5] | 10 |
| 0 | 1 | [1.3, 2.1, 0.3, 0.1, 1.4, 0.9] | 179 |
| 1 | 0 | [0.3, 0.5, 0.5, 1.4, 0.8, 1.4] | 32 |

PDNN modules are wrapped up into top-level python commands, which can be called directly in the bash scripts. All the configurations can be set in the bash scripts. For example, DNN training can be performed simply by:

```
python pdnn/run_DNN.py
  --train-data "train.pfile,partition=600m,random=true" \
  --valid-data "train.pfile,partition=600m,random=true" \
  --nnet-spec "250:1024:1024:1024:1024:1024:1901" \
  --output-format kaldi  --lrate "D:0.08:0.5:0.05,0.05:15" \
  --wdir exp/nnet  --output-file dnn.nnet
```

Details about the arguments of PDNN commands can be found on the website. These commands include:

- **run_SdA.py**. This command trains stacked denoising autoencoders (SdAs) [3], mostly used from DNN pre-training. A denoising autoencoder (DA) has the same structure as the traditional autoencoder, with the only difference of corrupting the input by adding some form of noise. SDAs are trained in a greedy layer-wise manner. Training of each DA involves reconstructing the clean input from the corrupted version of it.

- **run_RBM.py**. This command performs unsupervised training of the generative restricted Boltzmann machines (RBMs) model. A RBM is an undirected graphical model with a set of nodes representing visible units and a set of nodes representing hidden units. RBM training involves maximizing the likelihood of the observations with the contrastive divergence algorithm [4]. Similarly with SdAs, a stack of RBMs can be trained in a greedy layer-wise manner and used to initialize the parameters of DNNs. The first layer of a DNN corresponds to a Gaussian-Bernoulli RBM and each of the other hidden layers corresponds to a Bernoulli-Bernoulli RBM. Interested readers can refer to [5] for details on RBM.

- **run_DNN.py**. This command implements fine-tuning of DNNs based on stochastic gradient descent (SGD). Two special techniques can be adopted to improve DNN training.

  **Dropout:** Dropout has been proven to be effective in curbing overfitting and enhancing DNN training. In principal, on each presentation of a training example, dropout omits outputs from each hidden layer randomly via a binomial distribution. This distribution is governed by a pre-specified probability referred to as *dropout factor* in [6]. Dropout is applied only during training (fine-tuning). For testing (recognition), network parameters need to be scaled properly according to the value of the dropout factor [6]. In PDNN, dropout is triggered by the argument "--dropout-factor" to specify the value of the dropout factor. For the hidden layers, 0.2 has been found to be the optimal value for the dropout factor.

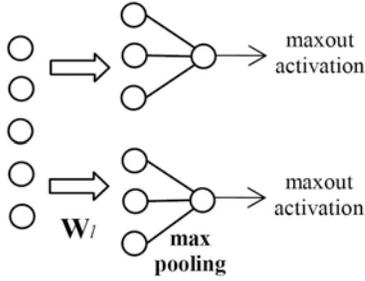

Figure 2. Maxout layer with the group size of 3.

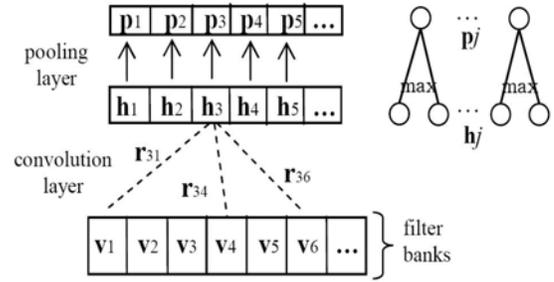

Figure 3. Illustration of the convolution layer and pooling layer.

**Maxout:** Previous works [7, 8, 9] have proposed to apply maxout networks to speech recognition. In maxout networks, the units at each hidden layer are divided into non-overlapping groups, each of which generates a single activation with max-pooling. Compared with standard DNNs, maxout networks result in superior performance on both hybrid systems and bottleneck-feature tandem systems. The advantages of maxout networks are two-fold. First, maxout networks can shrink the size of network parameters significantly, due to the reduced hidden activations. Second, maxout networks do not fix the shape of the activation function for hidden outputs. By tuning weight vectors of the subsumed hidden units, each maxout unit is capable of approximating any convex functions and thus can be optimized towards specific datasets in hand. Figure 2 illustrates the maxout layer in which the unit group size is 3.

- **run_CNN.py**. This command implements SGD-based fine-tuning of CNNs. CNNs can outperform DNNs on LVCSR tasks [10]. Instead of using the fully-connected parameter matrices, CNNs are characterized by parameter sharing and local feature filtering. On top of the convolution layer, a max-pooling layer is usually added to normalize spectral variations and reduce output dimensionality. Figure 3 shows our CNN architecture which works slightly different from the existing proposals [10]. In the convolution layer, we only consider filters along the frequency, assuming that the time variability can be modeled by HMM. Interested readers can refer to [11] for more details.

## 4. Recipes

Currently, Kaldi+PDNN contains four recipes. By default, they have the following configurations. In our descriptions, target_number means the number of classification targets at the final layer.

A. DNN Hybrid (run-dnn.sh)
  - The classic context-dependent DNN hybrid system
  - **Input**: spliced fMLLR features further projected by LDA to 250 dimensions
  - **Network**: 250:1024:1024:1024:1024:1024:target_ number

B. BNF Tandem (run-bnf-tandem.sh)
  - Fig. 4 shows the Deep Bottleneck Feature (DBNF) architecture [12] which our recipes are using to extract bottleneck front-end.
  - **Input**: spliced fMLLR features further projected by LDA to 250 dimensions
  - **Network**: 250:1024:1024:1024: 1024:1024:42:1024: target_number
  - LDA+MLLT and further MMI tandem systems are built on top of bottleneck features

C. BNF+DNN Hybrid (run-bnf-dnn.sh)
  - Hybrid system over spliced bottleneck features [13]
  - **DNN Hybrid Input**: spliced 9 frames of bottleneck features
  - **DNN Hybrid Network**: 378:1024:1024:1024:1024: target_ number

D. CNN Hybrid (run-cnn.sh)
  - Hybrid systems based on the deep convolution networks
  - **Input**: neighboring 11 frames of log-scale filter-bank features. These inputs are taken as 11 input feature maps.
  - **Network**: 2 convolution layers followed by 3 fully-connected layers
  - We apply 1-dimensional local filter (convolution kernel) only to the frequency axis. The max-pooling layer is inserted after each convolution layer.

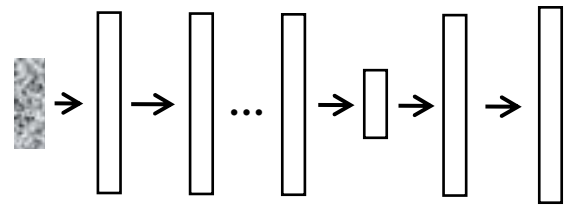

Figure 4. Architecture for the Deep Bottleneck Feature (DBNF) network [12].